\documentclass[runningheads]{llncs}
\usepackage[T1]{fontenc}
\usepackage{graphicx}

\usepackage{hyperref}
\usepackage{color}

\usepackage[caption=false]{subfig}

\usepackage{booktabs}

\usepackage[numbers]{natbib}

\begin{document}

\title{Explainable embeddings with Distance Explainer}

\author{Christiaan Meijer\inst{1}\orcidID{0000-0002-5529-5761} \and\\
E.~G.~Patrick Bos\inst{1,2}\orcidID{0000-0002-6033-960X}}

\authorrunning{Meijer and Bos}

\institute{Netherlands eScience Center, Science Park 402, 1098 XH Amsterdam, Netherlands,\\ \email{cwmeijer@protonmail.com}
\and
Center for Information Technology, University of Groningen, P.O. Box 11044, 9700 CA Groningen, Netherlands,\\ \email{e.g.p.bos@rug.nl}}

\maketitle

\begin{abstract}
    While eXplainable AI (XAI) has advanced significantly, few methods address interpretability in embedded vector spaces where dimensions represent complex abstractions. We introduce Distance Explainer, a novel method for generating local, post-hoc explanations of embedded spaces in machine learning models. Our approach adapts saliency-based techniques from RISE to explain the distance between two embedded data points by assigning attribution values through selective masking and distance-ranked mask filtering. We evaluate Distance Explainer on cross-modal embeddings (image-image and image-caption pairs) using established XAI metrics including Faithfulness, Sensitivity/Robustness, and Randomization. Experiments with ImageNet and CLIP models demonstrate that our method effectively identifies features contributing to similarity or dissimilarity between embedded data points while maintaining high robustness and consistency. We also explore how parameter tuning, particularly mask quantity and selection strategy, affects explanation quality. This work addresses a critical gap in XAI research and enhances transparency and trustworthiness in deep learning applications utilizing embedded spaces.

\end{abstract}

\section{Introduction}
Machine learning (ML) and deep learning (DL) methods have created high demand for understanding trained models, spurring the vibrant field of eXplainable AI (XAI).
While XAI algorithms are continuously developed for images, text, time-series and tabular data \citep{xaiclassification2021}, methods for general ``embedded spaces'' remain less common.

By ``embedded space'' we refer to a multi-dimensional vector space into which original data can be projected or encoded.
Embedded spaces are used in dimensional reduction operations in applications like FaceNet \citep{facenet}, Word2vec \citep{CHURCH_2017_word2vec}, VAE \citep{Kingma_2019_vae}, Spec2vec \citep{spec2vec2021}, and Life2Vec \citep{Savcisens2024life2vec}.
Multi-modal models such as CLIP \citep{clip2021} project multiple data modalities into shared embedded spaces.
The increasingly broad scientific application of embedded space models to model complex phenomena \citep[e.g.\ language acquisition research in][]{chrupala22} makes explainable methods especially promising for increasing research efficiency and trustworthiness \citep{yang+22,krishna+19,gevaert22}.

Embedded spaces created by deep neural nets are difficult to understand as their dimensions often represent multi-step abstractions \citep{Shahroudnejad2021ASO,zhu-individual-neuron-role}.
While some embedded spaces can be made more interpretable using XAI methods \citep{ali+23,boselli+24}, much work has focused on \emph{interpretability}\footnote{We follow the XAI nomenclature proposed by \citep{ali+23}.} of spaces or models as a whole \citep{boselli+24}, rather than \emph{explainability} of individual model decisions.
Recent work has adapted RISE to explain pairwise similarity in face recognition (S-RISE \citep{lu2023srise}) and face verification (CorrRISE \citep{lu2024corrrise}), using similarity-weighted mask aggregation for those domain-specific tasks.
However, general methods for explaining distances in arbitrary embedded spaces remain lacking.

Our method provides local, post-hoc explanations of distance between two data items' projections within embedded spaces.
This attribution approach \citep{Achtibat2023attribution} differs from methods like RISE \citep{petsiuk2018rise}, LIME \citep{ribeiro-etal-2016-trust} and GradCAM \citep{gradcam}, which take a single input, by instead comparing one data instance to a reference instance.
Unlike S-RISE and CorrRISE, our approach is modality-agnostic, applies to arbitrary embedded spaces, and introduces distance-ranked mask filtering with a mirror mode that replaces weighted summation.

In this paper we introduce a method to locally explain embedded spaces using attribution.
Section~\ref{sec:algorithm} describes our method.
We evaluate performance on models and data items listed in section~\ref{sec:experimental_setup} using quantitative measures (section~\ref{sec:quantitative_performance}) and qualitative assessment (section~\ref{sec:qualitative_assessment}).
Sections~\ref{sec:discussion} and~\ref{sec:conclusion} provide discussion and conclusions.

\section{Algorithm}
\label{sec:algorithm}

We consider a model that encodes data instances from multiple modalities into a single vector space.
For example, CLIP encodes both image and text data into a common semantic vector space, enabling reasoning about semantic similarity between different modalities.
Proximity in this space represents semantic similarity.
We aim to explain why certain instances end up closer to each other than others.

\subsection{Algorithm description}
\label{sec:algorithm_description}

Our starting point was RISE \citep{petsiuk2018rise}, which assigns saliency values to pixels by randomly masking them and examining the effect on the model's activation for a specified class.
Concretely, RISE generates many random binary masks, applies each to the input image (replacing masked pixels with a baseline value), forwards the masked image through the model, and computes a weighted average of masks using the resulting class scores as weights.
We chose RISE because an implementation was freely available, it is easy to reason about and extend, it is model-agnostic, and its random masking approach is sensitive to a wide range of semantic contents by considering combinations of parts rather than isolated regions.

\paragraph{Task}
The original RISE implementation \citep{petsiuk2018rise} targeted image classification.
In \cite{dianna}, RISE was extended to tabular data, text and time series.
We extended it to a task differing in two aspects:

\begin{enumerate}
\item Our task has different input and output types. Classification uses one input and outputs a class or activation vector. Our task takes two inputs and outputs a single distance between their embeddings.
\item RISE's saliency map equals a weighted sum over masks, with weights from the model's activation values \citep{petsiuk2018rise}. We lack class probabilities, instead having two embedded points with an associated distance. Converting this distance to a RISE weight was a crucial, non-trivial problem.
\end{enumerate}

These considerations led to three modifications:

\begin{enumerate}
    \item We define one input as ``reference'' and one as ``to-be-explained.'' The reference can be input in encoded form, implicitly supporting any modality. The to-be-explained item requires explicit masking strategy and visualization support per modality.
    
    \item We replace class-activation weights with cosine distance $d_\mathrm{cos}(e, r)$ between the to-be-explained item $e$ and reference $r$. However, we do not use $d_\mathrm{cos}(e, r)$ in weighted sums as in RISE (see Section~\ref{sec:ranking_or_weighting}).

    \item Instead, we introduce distance-ranked mask filtering, summing only masks meeting our filter criterion. Filtering proceeds via: (a) top $x\%$ of distances, (b) bottom $x\%$, or (c) ``mirror'' approach combining both.
    For low-distance masked images (b), masked pixels lacked salient information, contributing to highlighting minimally salient regions. High-distance masks (a) highlight maximally salient regions. After filtering, masks are summed without weights. In mirror mode, we subtract sets (a) and (b). The mirror method, assuming similar statistical properties of the two sets, improves signal-to-noise ratio through partial noise cancellation.
\end{enumerate}

\subsection{Considered algorithm alternatives}
\label{sec:considered_alternatives}

\subsubsection{Distance metric}
We chose cosine distance $d_\mathrm{cos}(x, y)$ over Euclidean distance because it emphasizes angular differences rather than vector magnitude.
This is crucial for our ImageNet classifier outputs (see section~\ref{sec:data_and_models}), where activation values sum to 1.
Euclidean distance depends on vector size, which ranges $[1/\sqrt{D},1]$ for dimension $D$, causing unwanted effects (smaller distances) when the model is uncertain.
For CLIP embeddings, different metrics might be suitable, but exploring this is beyond our scope.
In general, the ideal distance metric will depend on the particular embedded space properties.

\subsubsection{Ranking or weighting}
\label{sec:ranking_or_weighting}
We initially weighted masks proportional to $d_\mathrm{cos}(e, r)$, analogous to RISE.
However, in high-dimensional embedded spaces, distance differences between inputs were very small (typically $<10^{-4}$), making weights indistinguishable.
We attempted using $a^d$ (with $a \approx 20$) to amplify differences, but this required per-instance tuning of $a$, and often too few masks had significant weights, causing masking artifacts.
Our final approach ensures a fixed percentage of effective masks, avoiding artifacts.

\subsection{Algorithm summary}
The Distance Explainer algorithm proceeds as follows:

\begin{enumerate}
    \item The reference item $r$ is passed through the model to produce $\mathbf{x}_r$, used as fixed input.
    \item Repeat $N_\mathrm{masks}$ times on the to-be-explained item $e$:
    \begin{enumerate}
        \item Randomly mask $e$ via RISE (with parameters $c$) producing $M_i(e;c)$.
        \item Pass $M_i(e;c)$ through the model producing $\mathbf{x}_{M_i(e;c)}$.
        \item Calculate distance $d_i = d_\mathrm{cos}(M_i(e;c), r)$.
    \end{enumerate}
    \item Rank masked images using distances $\left\{d_i\right\}$.
    \item Apply selection filter on ranked masks.
    \item Sum remaining masks to produce the attribution map.
\end{enumerate}

Our implementation is available on GitHub\footnote{\url{https://github.com/dianna-ai/distance_explainer}}.

\section{Experimental setup}
\label{sec:experimental_setup}

We describe the models and data items used to assess our explainer in sections~\ref{sec:quantitative_performance} and~\ref{sec:qualitative_assessment}.
A gallery of all results is on Zenodo\footnote{\label{foot:zenodo_gallery}\url{https://zenodo.org/records/14044386}} and code on GitHub\footnote{\url{https://github.com/dianna-ai/explainable_embedding/}}.

\subsection{Data item modalities and embedded space models}
\label{sec:data_and_models}

We experimented with two modality pairs:

\begin{description}
    \item[Image vs image] Using ImageNet models to produce 1000-dimensional classification vectors. While not typically considered embedded space vectors, they can be interpreted as semantic vectors in a well-defined space, making them ideal for experiments where we can unequivocally interpret vectors. We primarily used ResNet50 from Keras; in section~\ref{sec:randomization} we used VGG16 for its fewer layers.
    \item[Image vs caption] Using ViT-B/32 \citep{Dosovitskiy2020AnII} CLIP model \citep{clip2021}, transforming both images and captions to a common 512-dimensional semantic space.
\end{description}

We restrict ourselves to attribution maps on images, using captions only as reference items\footnote{Other domains require defining masking functions.
DIANNA \citep{dianna} implements masking for text, tables and time-series \citep{meijer2024maskingtimeseries}.}.

\subsection{Data items}
\label{sec:data_items}
For ImageNet, data pairs probe four application areas: \textbf{Same class} (two bee images), \textbf{Multiple classes per image} (dog and car vs. single objects), \textbf{Close/related classes} (bee vs. fly; car vs. bike), and \textbf{Unrelated items} (flower vs. car).
We define ``related'' as sharing a superordinate category in the ImageNet hierarchy (e.g.\ both insects), and ``unrelated'' as lacking such overlap; these assignments were made by the authors based on intuitive semantic relatedness.

For CLIP, we tested: bee image with captions ``a bee sitting on a flower'', ``a bee'', ``an image of a bee'', ``a fly'', and ``a flower''; labradoodle with ``a labradoodle''; dog-and-car image with ``a car'' and ``a dog''; flower with ``a car''; car with ``a bicycle.''

\section{Quantitative performance results}
\label{sec:quantitative_performance}

We evaluate quantitative performance aspects of our explainer.
Following \cite{nauta2023}, we assess Correctness and Completeness via Incremental Deletion \citep{willemthesis} and MPRT \citep{adebayo2018mprt}, and Continuity via Average Sensitivity \citep{bhatt2020evaluating}.
While \cite{hedström2024sanity-revisited} provides improved MPRT robustness, we qualitatively interpret attribution maps from intermediate MPRT steps rather than using scores.
We used Quantus \citep{quantus2023} implementations adapted for our task.
Our terminology (faithfulness, sensitivity/robustness, randomization) aligns with Nauta's (Correctness, Continuity, Completeness).

\subsection{Faithfulness}

\emph{Faithfulness} measures whether altering highlighted regions produces corresponding output changes.
A faithful explanation identifies elements genuinely influencing the model's decisions.

\begin{figure}[!htbp]
    \centering
    \includegraphics[width=\textwidth]{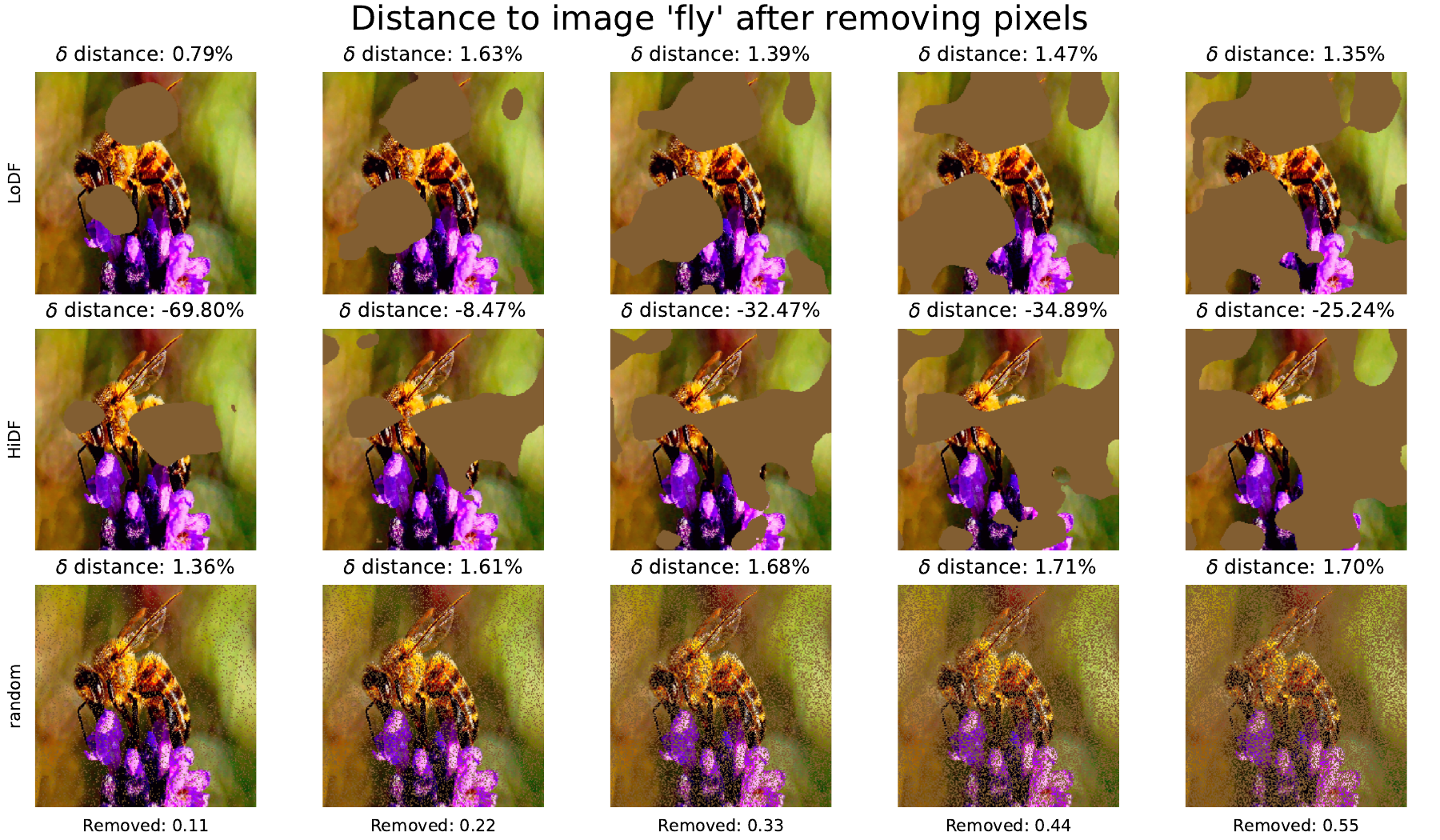}
    \caption{Incremental deletion on the bee image whose distance to \textit{a fly}'s image is explained. Deleted pixels are brown. Left to right shows increasing deletion. Top: LoDF. Middle: HiDF. Bottom: random.}
    \label{fig:incremental_deletion_bee_vs_fly_pixel_removal}
\end{figure}

\begin{figure}[!htbp]
    \centering
    \subfloat[LoDF vs random.\label{fig:incremental_deletion_bee_vs_fly_graphs_LoDF}]{%
        \includegraphics[width=0.49\textwidth]{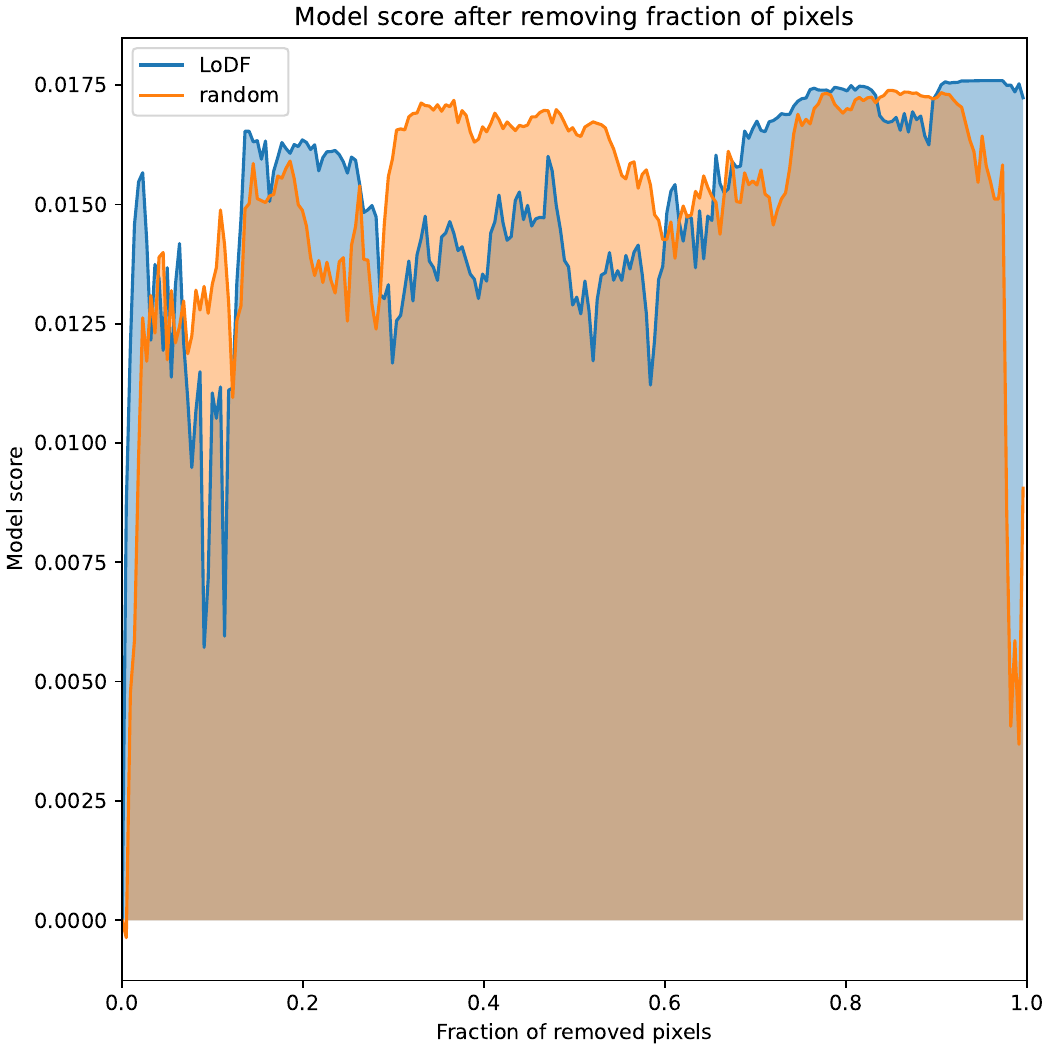}%
    }
    \hfill
    \subfloat[HiDF vs random.\label{fig:incremental_deletion_bee_vs_fly_graphs_HiDF}]{%
        \includegraphics[width=0.49\textwidth]{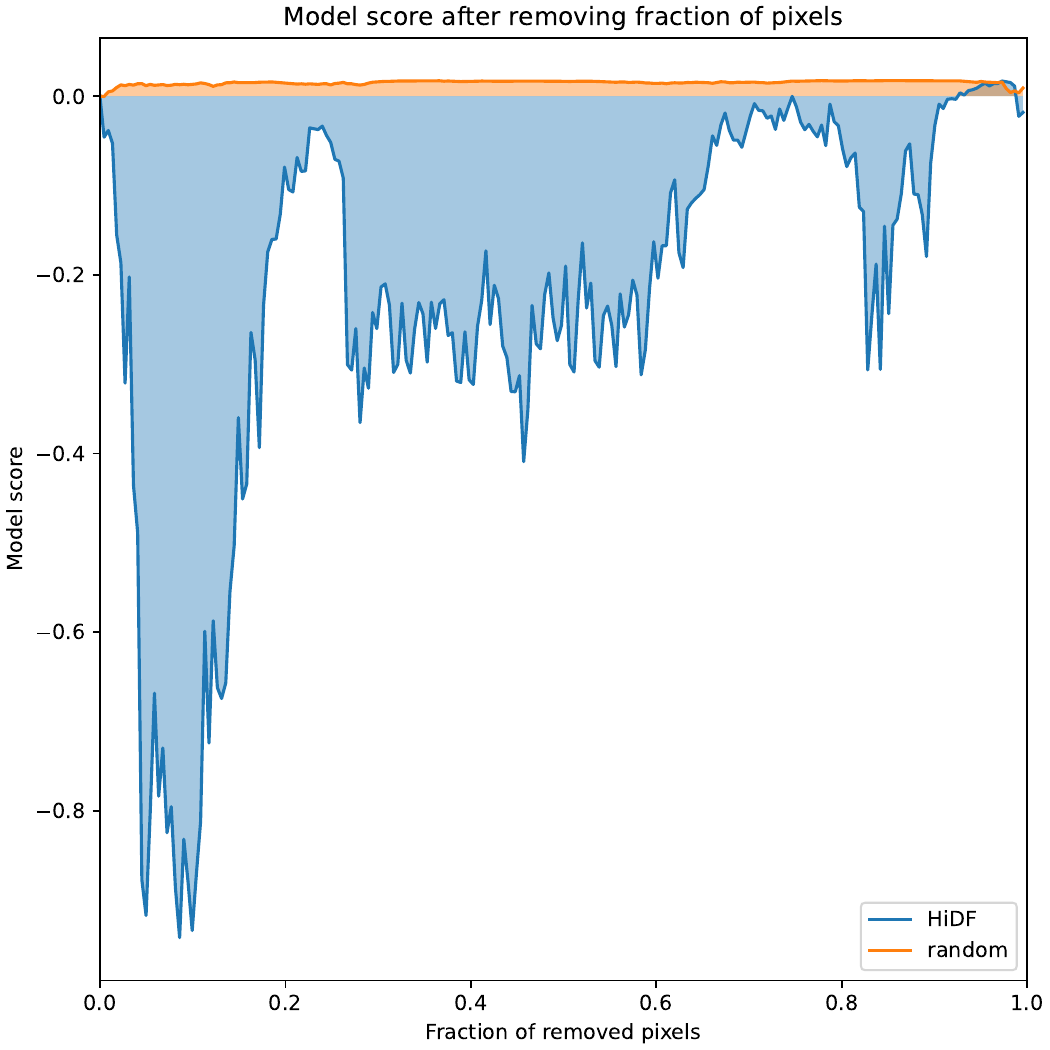}%
    }
    \caption{Distance change ($\Delta d = d_\mathrm{cos}(\mathrm{deleted}, r) - d_\mathrm{cos}(e, r)$) under incremental deletion on bee vs.\ fly. Vertical: $\Delta d$ (negative means the deletion reduced distance to the reference). Horizontal: deleted pixel percentage.}
    \label{fig:incremental_deletion_bee_vs_fly_graphs}
\end{figure}

We measured faithfulness via incremental deletion with three orders: (1) LoDF (low distance first, i.e. start deletion with pixels that contribute most to the data items having a low distance), (2) HiDF (high distance first)\footnote{We avoid ``most/least relevant first'' (MoRF/LeRF) terminology because our bidirectional maps show contributions to both increased and decreased distance.}, (3) random.
Results for bee vs. fly appear in Figures~\ref{fig:incremental_deletion_bee_vs_fly_pixel_removal} and~\ref{fig:incremental_deletion_bee_vs_fly_graphs}.

Figure~\ref{fig:incremental_deletion_bee_vs_fly_graphs_HiDF} removes least fly-like pixels first --- pixels indicative of bees and not flies.
Since the model classifies the original as bee, removing these pixels causes large negative $\Delta d$ values.
Figure~\ref{fig:incremental_deletion_bee_vs_fly_graphs_LoDF} removes most fly-like pixels first.
These barely affect bee classification scores since the model already confidently predicts bee, resulting in small changes comparable to random removal.

\subsection{Sensitivity / robustness}
\label{sec:sensitivity_robustness}
Robustness means small input changes produce correspondingly small output changes, ensuring stable, non-oversensitive results.

We measured robustness via Average Sensitivity \citep{bhatt2020evaluating} using Quantus \citep{quantus2023}, with deterministic RISE mask generation to avoid dual randomness sources.
Parameters: \texttt{nr\_samples} = 20, \texttt{perturb\_std} = $0.1 \times 255$, 500 masks.
Results: 0.06 for bee vs. fly, 0.04 for bee vs. bee --- low sensitivity indicating high robustness (typical sensitivity values range 0--1 \citep{yeh2019avgsensitivity}).

\subsection{Randomization}
\label{sec:randomization}

A model agnostic XAI method's output should depend highly on the model.
In pathological cases, methods may themselves encode prior knowledge (e.g.~edge detection in images), producing similar explanations regardless of model parameters.
\emph{Randomization} tests assess whether explainer output changes appropriately when model parameters are randomly changed.

\begin{figure}[htbp!]
    \centering
    \includegraphics[width=1\linewidth]{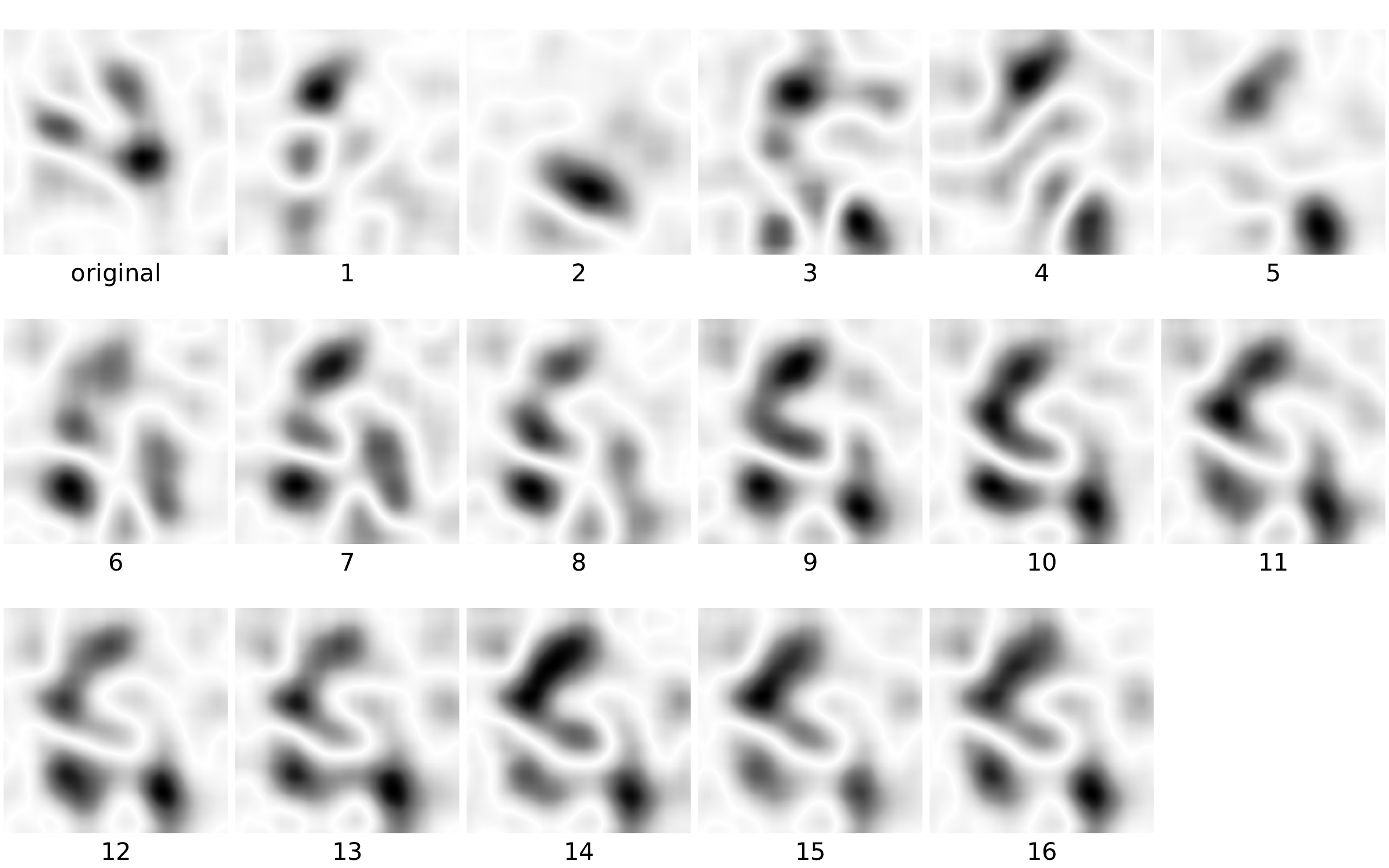}
    \caption{MPRT (top-down): First image shows the unperturbed attribution map. Subsequent images show maps with iteratively one additional layer's weights randomized, starting with the final layer. Loss of structure indicates dependence on learned parameters.}
    \label{fig:mprt_topdown}
\end{figure}

\begin{figure}[htbp!]
    \centering
    \includegraphics[width=1\linewidth]{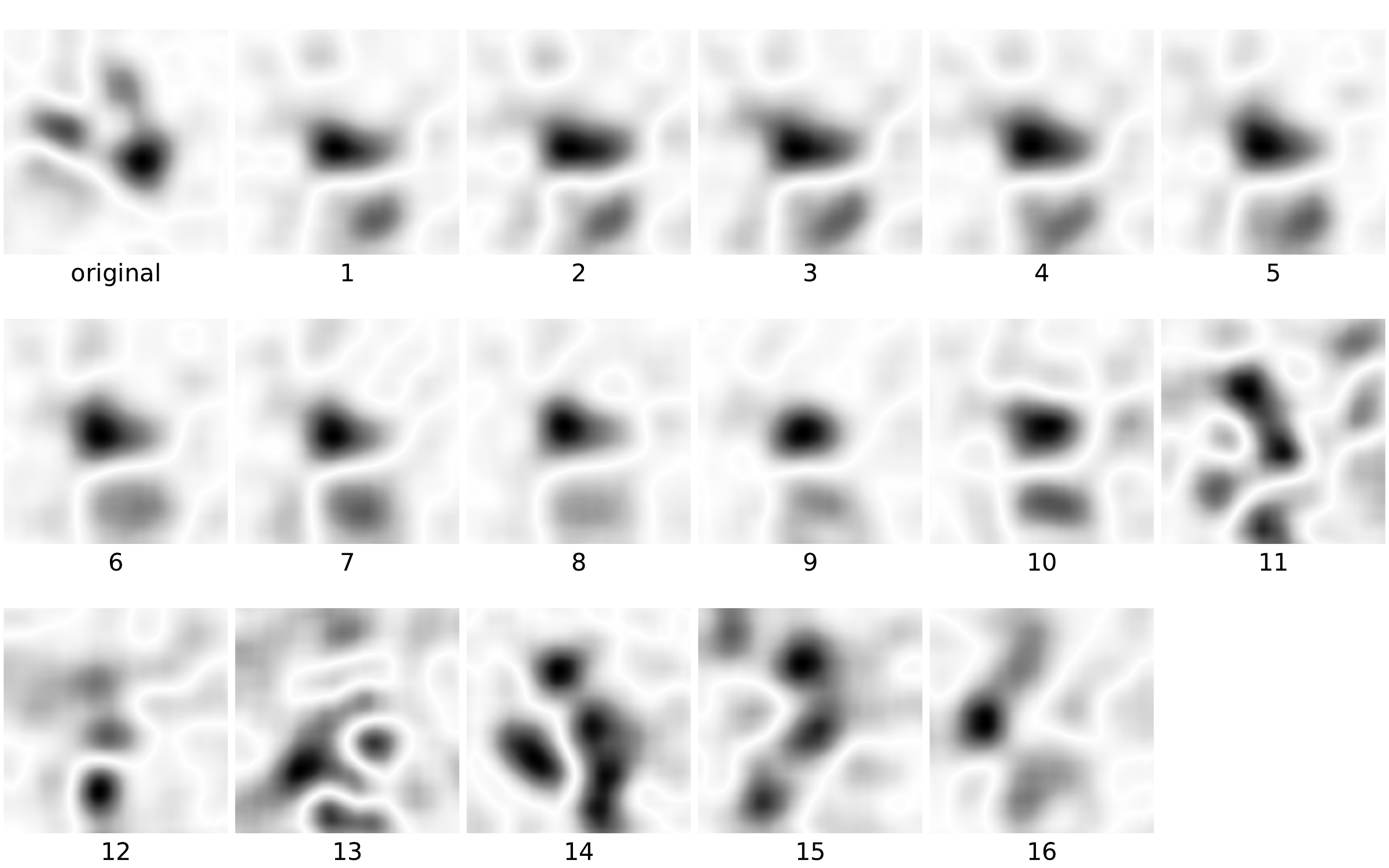}
    \caption{MPRT (bottom-up): Like Figure~\ref{fig:mprt_topdown}, but perturbation starts with the first layer.}
    \label{fig:mprt_bottomup}
\end{figure}

We used Model Parameter Randomization Test (MPRT) \citep{adebayo2018mprt}, which shuffles layer weights.
Quantus offers three modes: top-down, bottom-up, and independent shuffling.
MPRT produces Spearman correlations between original and perturbed attribution maps.
Table~\ref{tab:mprt} shows scores; Figures~\ref{fig:mprt_topdown} and~\ref{fig:mprt_bottomup} show attribution maps\footnote{We modified Quantus code to store intermediate maps.} for bee vs. fly with 1000 masks.

\begin{table}[htbp!]
\centering
\begin{tabular}{lrrr}
\toprule
Layer & Top down & Bottom up & Independent \\
\midrule
1 & 0.09 & 0.30 & 0.11 \\
2 & 0.28 & 0.05 & 0.10 \\
3 & -0.10 & 0.22 & 0.11 \\
4 & 0.02 & -0.18 & 0.92 \\
5 & -0.24 & -0.09 & 0.51 \\
6 & -0.11 & 0.17 & 0.38 \\
7 & -0.09 & 0.01 & 0.67 \\
8 & -0.04 & -0.02 & 0.61 \\
9 & -0.11 & 0.02 & 0.32 \\
10 & -0.15 & 0.03 & 0.60 \\
11 & -0.13 & -0.03 & 0.56 \\
12 & -0.19 & 0.02 & 0.63 \\
13 & -0.19 & 0.01 & 0.41 \\
14 & -0.21 & -0.03 & 0.42 \\
15 & -0.27 & -0.01 & 0.54 \\
16 & -0.22 & 0.02 & 0.18 \\
\bottomrule
\end{tabular}
\caption{MPRT scores between 1 (perfect positive correlation) and -1 (perfect negative).}
\label{tab:mprt}
\end{table}

Desirable behavior: attribution maps change completely when randomizing layers, indicating high model dependency without invalid assumptions.
Figures~\ref{fig:mprt_topdown} and~\ref{fig:mprt_bottomup} demonstrate this.
In both top-down and bottom-up randomization, structures are lost immediately after randomizing the first layer, confirming the explainer's dependence on learned parameters.
All modes also show low correlations in Table~\ref{tab:mprt} after first-layer randomization, indicating high attribution map dependency on the model.

\section{Qualitative assessment}
\label{sec:qualitative_assessment}

We complement quantitative assessment with visual inspection and qualitative findings, covering aspects like Nauta's Consistency \citep{nauta2023} (section~\ref{sec:number_of_masks}), Contrastivity (``Unrelated'' items in section~\ref{sec:data_items}), and Coherence.

\subsection{Resulting attribution maps using default parameters}
\label{sec:look_it_works}

We present attribution maps for all data pairs (section~\ref{sec:data_items}) using default settings (motivated in sections~\ref{sec:parameter_exploration} and~\ref{sec:mask_selection}).

\begin{figure}[htbp!]
    \centering
    \includegraphics[width=\textwidth]{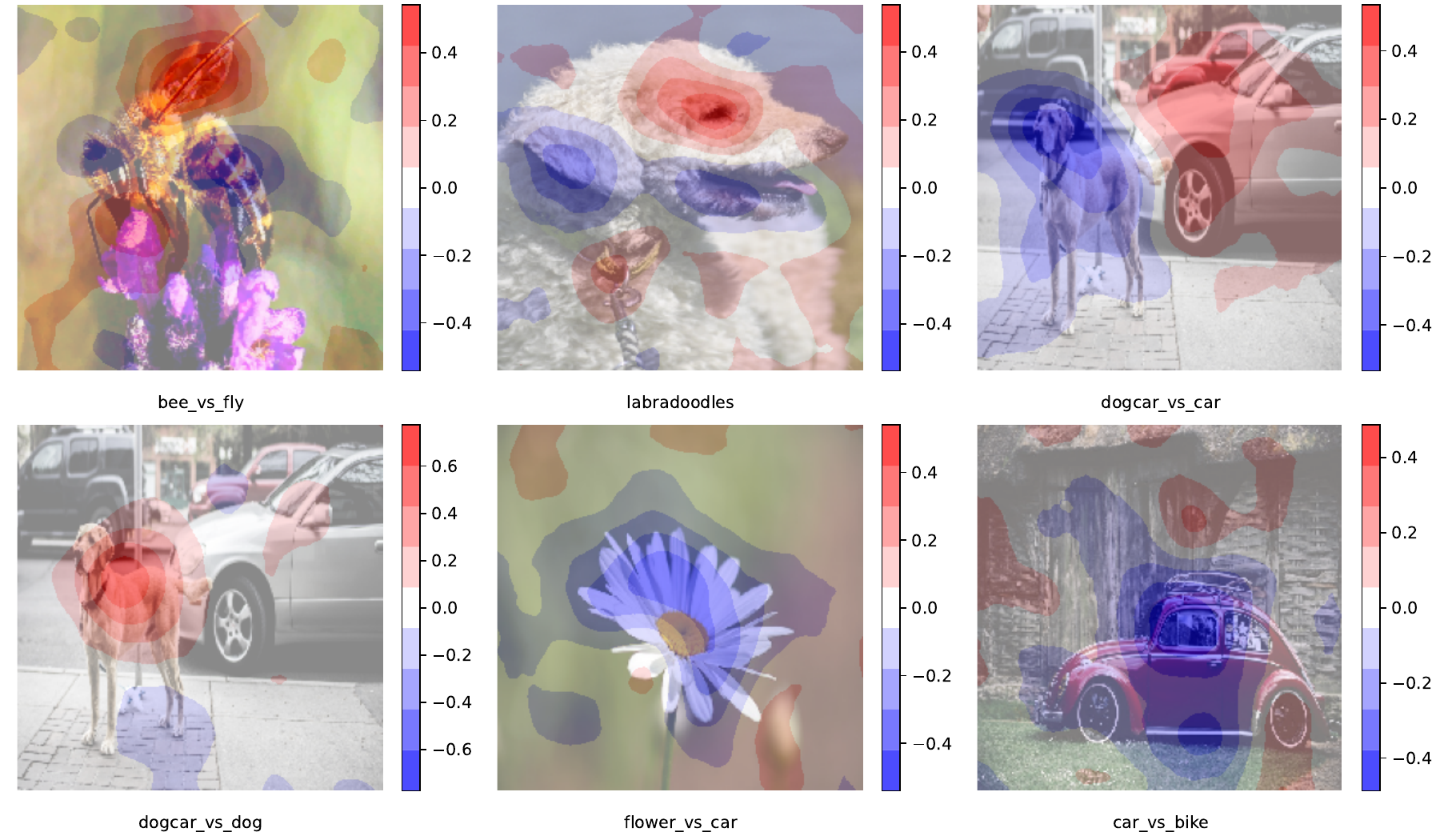}
    \caption{Attribution maps using default parameters on image-versus-image pairs. Red shades indicate regions that decrease distance to the reference; blue shades indicate regions that increase distance. Intensity reflects attribution magnitude.}
    \label{fig:item-pair_image_vs_image_gallery}
\end{figure}

Figure~\ref{fig:item-pair_image_vs_image_gallery} shows image-versus-image results.
Bee vs. fly: wings bring images closer; stripes drive them apart.
Labradoodle: eyes and collar are distinctive.
Dog-and-car: car highlighted when reference is car; dog highlighted when reference is dog.
Flower vs. car: flower moves away from car region with no closer regions.
Car vs. bicycle: car differs from bicycle, but wheels are excluded from strongly differing areas --- bicycles share wheels with cars.

\begin{figure}[htbp!]
    \centering
    \includegraphics[width=\textwidth]{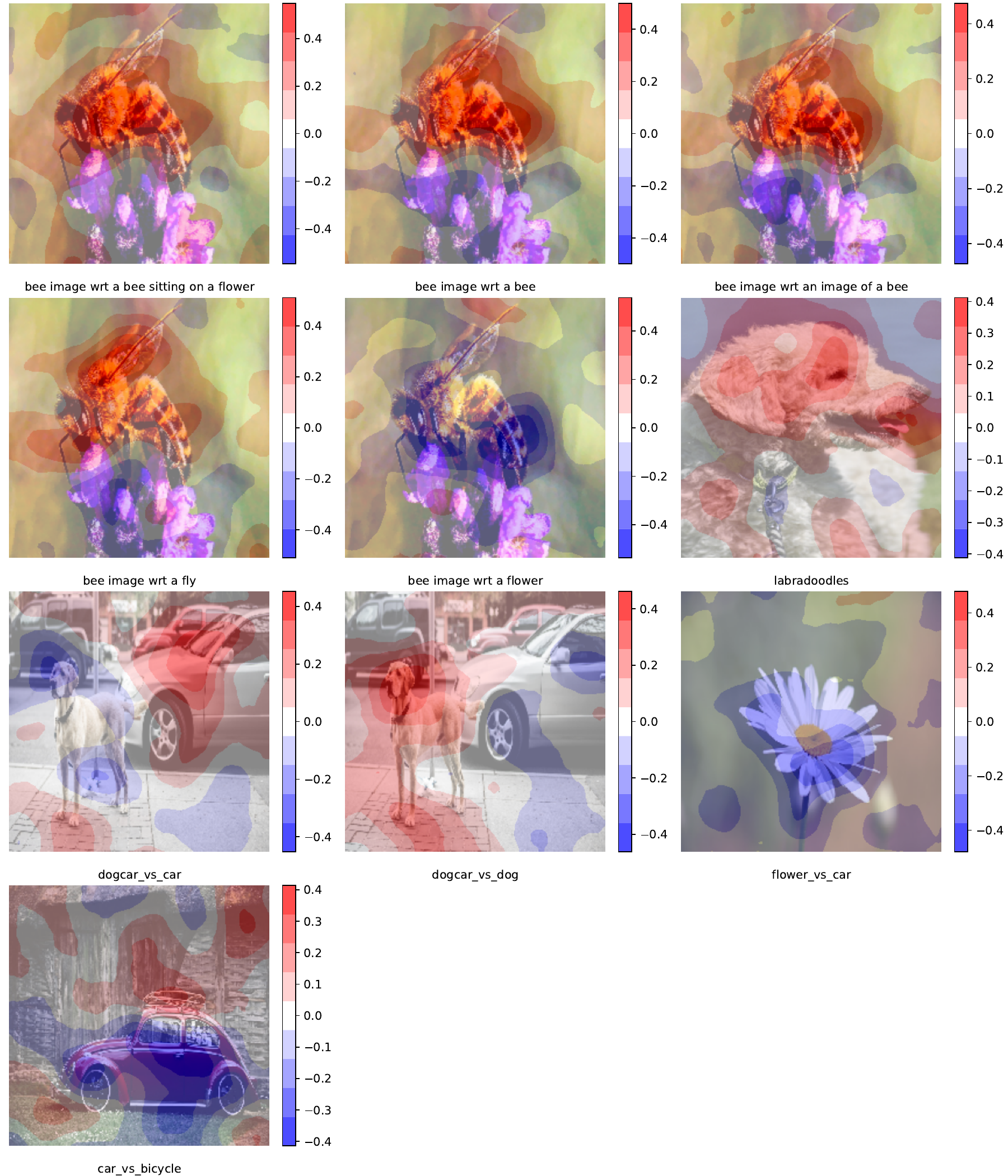}
    \caption{Attribution maps using default parameters on image-versus-caption pairs. Color scale as in Figure~\ref{fig:item-pair_image_vs_image_gallery}.}
    \label{fig:item-pair_image_vs_caption_gallery}
\end{figure}

Figure~\ref{fig:item-pair_image_vs_caption_gallery} shows image-versus-caption results.
These attributions seem slightly less sharp, possibly due to the model or suboptimal parameters, but performance remains convincing.

\subsection{Parameter exploration}
\label{sec:parameter_exploration}

While we minimized free parameters, we found default values giving decent results for our experiments.
Different situations may require tuning.
Table~\ref{tab:defaults} lists the defaults used throughout unless noted otherwise; each is motivated in the subsections below.
We explore non-default choices as a starting point for optimization beyond this work.

\begin{table}[htbp!]
\centering
\begin{tabular}{llr}
\toprule
Parameter & Symbol & Default \\
\midrule
Number of masks & $N_\mathrm{masks}$ & 1000 \\
Mask coverage percentage & $p_\mathrm{keep}$ & 0.5 \\
Mask feature resolution & & $8\times8$ \\
Selection mode & & mirror (two-sided) \\
Selection threshold & & 10\% per side \\
\bottomrule
\end{tabular}
\caption{Default hyperparameter values used in all experiments unless stated otherwise.}
\label{tab:defaults}
\end{table}

\subsubsection{Number of masks}
\label{sec:number_of_masks}

\begin{figure}[htbp!]
    \centering
    \includegraphics[width=\textwidth]{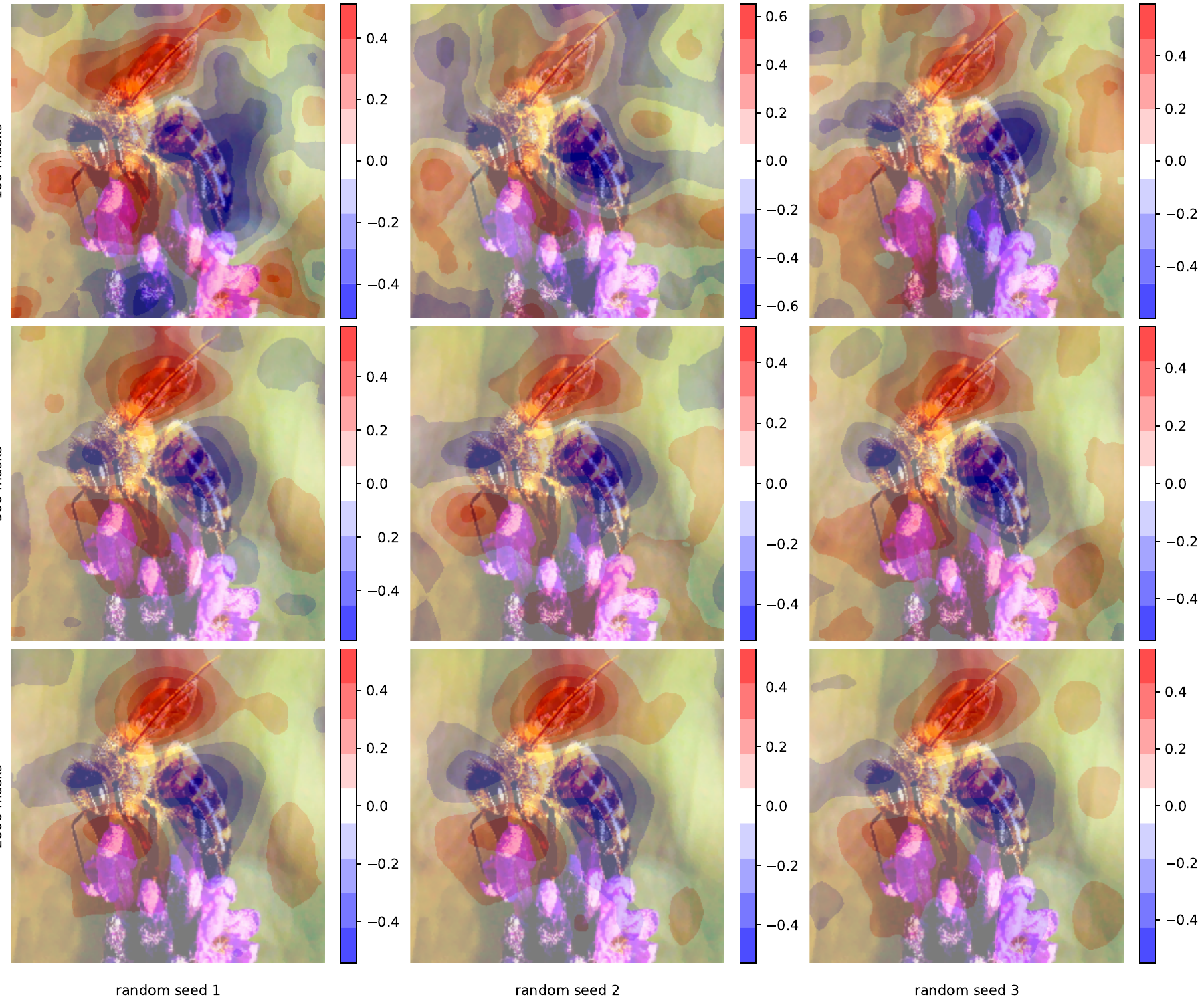}
    \caption{Attribution map convergence with increasing number of masks (rows) for three different random seeds (columns), bee vs.\ fly case. Patterns become increasingly similar across seeds as mask count grows.}
    \label{fig:number_of_masks_convergence}
\end{figure}

More masks increase stability through more samples.
Table~\ref{tab:number_of_masks_convergence} shows decreasing differences between maps with different random masks as mask count increases.
Figure~\ref{fig:number_of_masks_convergence} confirms patterns become increasingly similar.

\begin{table}[htbp!]
\centering
\begin{tabular}{rrrr}
\toprule
Number of masks & \multicolumn{3}{r}{Mean STD per pixel} \\
 & bee vs fly & flower vs car & dogcar vs car \\
\midrule
100 & 0.139 & 0.140 & 0.140 \\
500 & 0.064 & 0.062 & 0.062 \\
2000 & 0.031 & 0.031 & 0.032 \\
\bottomrule
\end{tabular}
\caption{Increasing masks decreases standard deviation between maps with different random seeds, regardless of data pairs.}
\label{tab:number_of_masks_convergence}
\end{table}

These numbers do not generalize to all Distance Explainer use-cases.
Different images, models, or parameters may require fine-tuning.
Unless noted, we use 1000 masks, providing decent trade-offs between stability, cost, and complexity.

\subsubsection{Mask coverage percentage}

The proportion of pixels to keep unmasked in each random mask, $p_\mathrm{keep}$, plays a subtle role.
The optimal value depends on the specific data item and salient features.

\begin{figure}[!htbp]
    \centering
    \includegraphics[width=\textwidth]{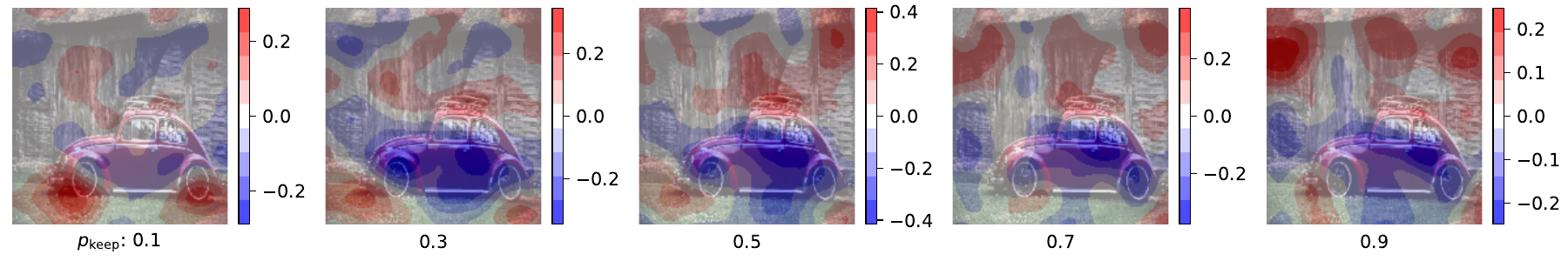}
    \caption{Image of a car versus caption ``a bicycle''. From left to right, $p_\mathrm{keep}$ is increased, showing values: 0.1, 0.3, 0.5, 0.7, 0.9.}
    \label{fig:parameter exploration p_keep car bicycle}
\end{figure}

For the car versus bicycle caption case (Figure~\ref{fig:parameter exploration p_keep car bicycle}), two interesting $p_\mathrm{keep}$ ranges emerge: values of 0.01--0.1 highlight wheels (when only wheels remain, the model may interpret them as bicycle parts), while 0.2--0.9 highlight the car body.
This illustrates that data item particulars significantly affect performance under different parameter settings.

For most cases (like the bee image versus caption one), the explainer works best at central $p_\mathrm{keep}$ values (0.4--0.6).
Performance decreases toward 0 or 1, manifesting as noisy patches of high attribution values that don't reflect salient image parts.
The average amplitude also decreases at extremes, increasing relative noise.
Additional examples of parameter sweeps across various image/caption pairs are available in our Zenodo gallery (see footnote \ref{foot:zenodo_gallery}).

We recommend users sweep this parameter.
Multi-parameter combination of attribution maps could be explored in future work.

\subsubsection{Mask feature resolution}
\label{sec:mask feature resolution}

\begin{figure}[htbp!]
    \centering
    \includegraphics[width=\textwidth]{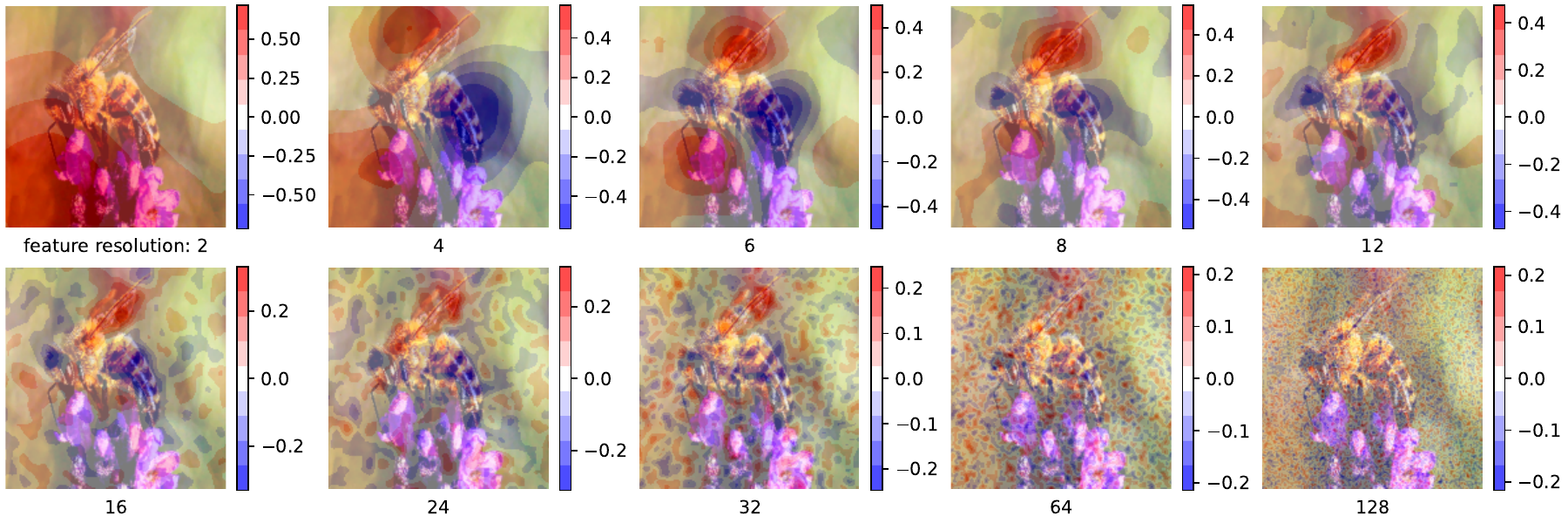}
    \caption{The effect of using different mask feature resolution is shown. The number of features in both image axis directions is shown below each panel.}
    \label{fig:feature_res_increase}
\end{figure}

The number of ``superpixel'' areas for masking affects explanation quality.
Figure~\ref{fig:feature_res_increase} shows that resolutions of 2 and 4 are too coarse, while values above 32 become too noisy (though this can be compensated with more masks).
At 8--16, salient parts like bee wings are more precisely delineated.
When differentiating similar classes, higher feature resolution is necessary to capture details.
The optimal value depends on the explanation's purpose and a cost-benefit analysis, as finer resolution requires more masks.

We confirmed that increasing mask numbers reduces noise at higher resolutions.
For image-versus-caption cases, even more masks are needed to reach the same noise reduction level as image-versus-image cases, consistent with our earlier observations in section~\ref{sec:look_it_works}.

\subsection{Mask selection}
\label{sec:mask_selection}

We replaced RISE's mask \emph{weighting} with mask \emph{selection}.
We explored multiple selection methods.

\subsubsection{One-sided vs two-sided}

One-sided selection uses only top $n$-percent masks (decreasing distance).
Two-sided uses both top and bottom $n$-percent (also increasing distance), multiplying bottom masks by -1 before adding.

\begin{figure}[htbp!]
    \centering
    \includegraphics[width=\textwidth]{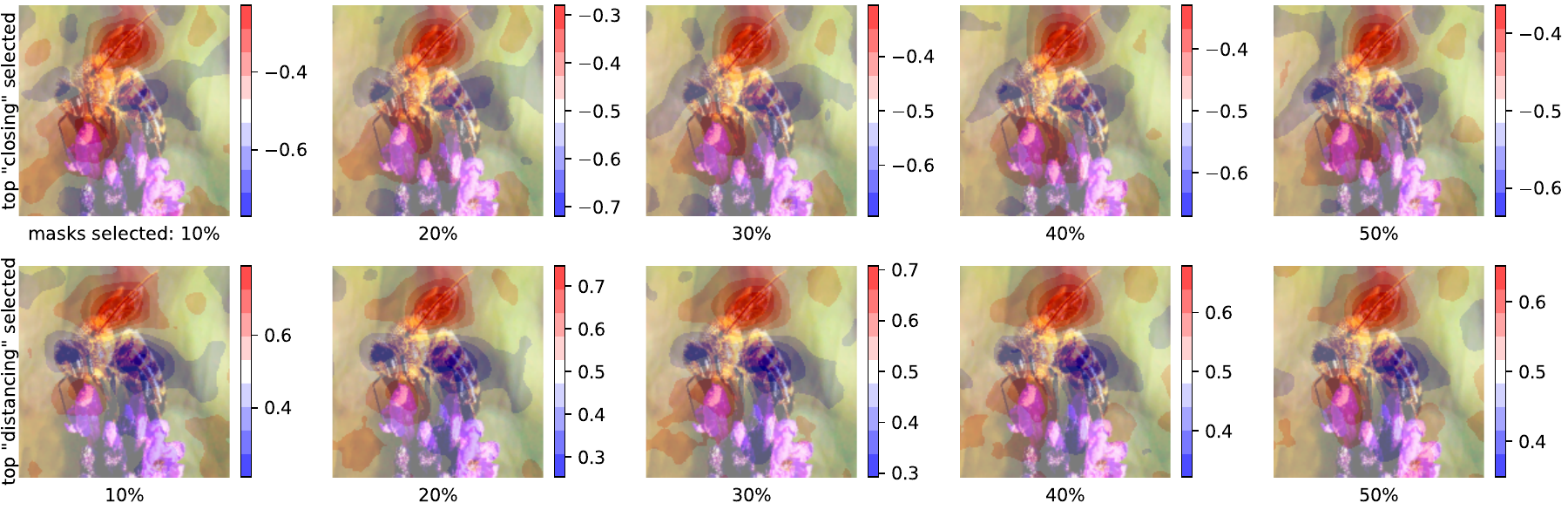}
    \caption{One-sided mask selection on bee vs.\ fly. Top row: selecting only distance-decreasing masks. Bottom row: selecting only distance-increasing masks (multiplied by $-1$ for comparison). Left to right: increasing percentage of selected masks.}
    \label{fig:masking_onesided}
\end{figure}

Figure~\ref{fig:masking_onesided} compares bee vs. fly using top and bottom $n$-percent masks.
Both yield nearly indistinguishable patterns, measuring similar signals.
We conclude that the two-sided approach is better, using twice as many masks, averaging out more noise.
Across 10-50\% selection, signals remain stable, suggesting either most amplitude is in the first 10\% or all ranges contain similar information and noise.

\subsubsection{Threshold value}
\begin{figure}[htbp!]
    \centering
    \includegraphics[width=\textwidth]{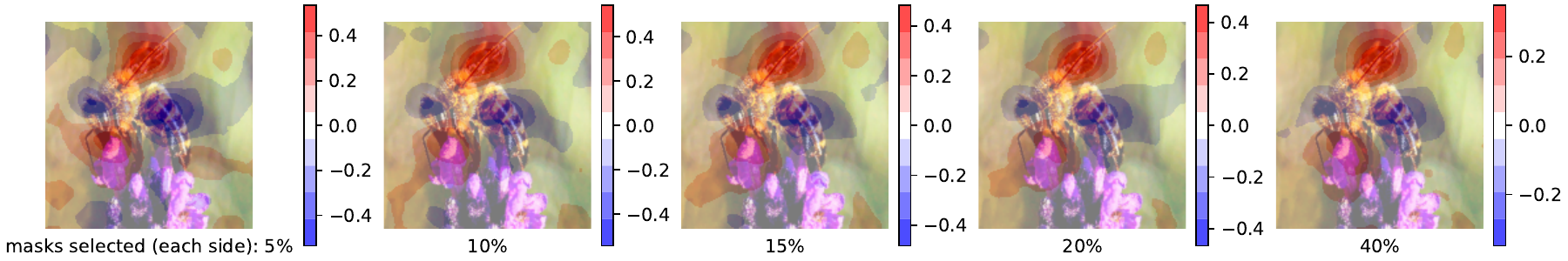}
    \caption{Two-sided ``mirror'' selection on bee vs.\ fly. Left to right: increasing percentage of included masks. Color scale as in Figure~\ref{fig:item-pair_image_vs_image_gallery}.}
    \label{fig:masking_threshold}
\end{figure}

Using two-sided ``mirror'' mode, Figure~\ref{fig:masking_threshold} shows increasing selection thresholds produces no meaningful changes beyond 10\%.
Neither highlighted regions nor map quality (e.g. noise patterns) change significantly.
We confirmed visually that masks in selection ranges below the top 10\% give higher noise levels.
We set default threshold to 10\%.

\section{Discussion}
\label{sec:discussion}

Our method provides a novel approach to explaining distances in embedded spaces via saliency-based attribution.
Experiments demonstrate effective identification of features contributing to similarity or dissimilarity between embedded data points.
While our experiments focus on image-based embeddings, the methodology is not inherently limited to images.
The algorithm operates on masked inputs and embedding distances, requiring only a modality-specific masking function.
DIANNA \citep{dianna} already provides such masking for text, tabular data and time-series \citep{meijer2024maskingtimeseries}, making extension to these modalities straightforward.
Our image-versus-caption experiments with CLIP demonstrate that the method works across modalities with comparable results.
To support reproducibility, we openly provide the method implementation, all experiment code, and result data (see Section~\ref{sec:experimental_setup}).

Computational costs are dominated by model runs for masked outputs, approximately equal to model cost multiplied by the number of masks.
Random mask creation is non-trivial but, with our parameter choices, substantially cheaper than model runs for images, though costs may vary across data modalities.
Overall, computational requirements are similar to those of RISE.

Key challenges include parameter tuning, particularly mask numbers and filtering criteria.
Random masking introduces trade-offs between computational cost and stability.
Different embedding models may require tailored distance metrics.
Alternative perturbation strategies could reduce out-of-distribution (OOD) input risks.
RISE replaces masked regions with a fixed baseline (e.g.\ black pixels), which can produce inputs far from the training distribution; this may lead model responses to reflect OOD artifacts rather than genuine feature importance.
We left this imputation strategy unchanged in our algorithm, but investigating learned or blurred infilling as alternatives is worthwhile.
Similarly, guided or stratified mask sampling could improve efficiency by concentrating perturbations on informative regions, reducing the number of masks needed for convergence.
We evaluated only RISE as attribution backend; comparing alternative perturbation or gradient-based engines is a natural next step.
Systematic hyperparameter sensitivity analyses and formal attribution uncertainty estimates are likewise warranted.

While attribution maps provide meaningful insights for AI researchers and developers, their utility for non-experts remains open.
Future work could explore user studies assessing explanation comprehensibility across audiences, particularly how academic researchers can leverage this tool to improve AI-enhanced research.

\section{Conclusion}
\label{sec:conclusion}

We introduced a method explaining distances in embedded spaces using a saliency-based approach adapted from RISE.
By analyzing input perturbation impacts on similarity metrics, our method generates local explanations highlighting features contributing most to embedding proximity or separation.

Experimental results demonstrate efficacy across different models and data modalities, particularly in image-based and multi-modal embeddings.
Quantitative evaluations confirm our method maintains robustness, consistency, and dependency on model parameters, aligning with established XAI evaluation criteria.

Future work could refine the method for text and other non-visual embeddings, explore alternative distance metrics, and optimize or automate parameter selection.
Studies on human interpretability could provide insights into usability in real-world applications.

\begin{credits}
\subsubsection{\ackname} We thank Willem van der Spek for fruitful discussions and feedback on quantitative evaluation and for providing an Incremental Deletion implementation.
We also thank Jisk Attema and Elena Ranguelova for helpful discussions about the algorithm and experimental setup.
We thank the anonymous reviewers for their constructive feedback, which improved the clarity and completeness of the paper.
Finally, we express our deep gratitude to Vikram Sunkara for presenting our work at the XAI-2026 conference and for his insightful comments on the presentation.
Experiments were run on the DAS-6 cluster \citep{das}.

\subsubsection{Research software usage}
Software used in our Distance Explainer algorithm includes: DIANNA \citep{dianna}, NumPy \citep{numpy}, scikit-learn \citep{scikit-learn}, tqdm, pyyaml and dataclass\_wizard.
For analysis, we additionally used Matplotlib \citep{matplotlib}, Quantus \citep{quantus2023}, Keras \citep{keras}, PyTorch \citep{pytorch}, torchtext, CLIP \citep{clip2021}, gitpython and Pillow.

\subsubsection{Generative AI assistance}
We used ChatGPT 4o on 21 Jan 2025 for refining the Discussion and Conclusion texts, used NotebookLM on 28 Jan 2025 to refine our introduction text, again ChatGPT 4o through Copilot on 11, 24 and 25 April 2025 for refining sections~\ref{sec:quantitative_performance} and~\ref{sec:qualitative_assessment} and Claude 3.7 Sonnet and Mistral on 6 May 2025 to refine section~\ref{sec:quantitative_performance}.
The abstract was written using Claude 3.7 Sonnet and NotebookLM on 20 May 2025.
For XAI 2026 submission we used Claude 4.5 Sonnet to rewrite more concisely.
For camera-ready revisions on 13 March 2026 we used Claude Opus 4.6 via GitHub Copilot to address reviewer comments.
All AI-output has been verified for correctness, accuracy and completeness, adapted where needed, and approved by the authors.

\subsubsection{\discintname}
The authors have no competing interests to declare that are relevant to the content of this article.

\end{credits}

\bibliographystyle{splncs04}
\bibliography{bibliography}

\end{document}